\documentclass[letterpaper, 10 pt, conference]{ieeeconf}  

\IEEEoverridecommandlockouts                              

\title{\LARGE \bf
Higher or Lower: Challenges in Object based SLAM
}

\author{Zhihe Zhang*, Hao Wei, Hongtao Nie
\thanks{Zhihe Zhang*, Hao Wei, Hongtao Nie are with the School of Information Science \& Engineering, Lanzhou University, Lanzhou 730000, China. E-mail:
        {\tt\small zzhihe107@gmail.com        }}%
}

\usepackage{algorithmicx}
\usepackage{algorithm}
\usepackage{amssymb}
\usepackage{amsmath}
\usepackage{algpseudocode}
\usepackage{booktabs}
\usepackage[table,xcdraw]{xcolor}
\usepackage{graphicx}
\usepackage{stfloats}
\usepackage{threeparttable}
\usepackage{cite}
\usepackage{multirow}
\usepackage{tabularx}
\begin{document}

\maketitle
\thispagestyle{empty}
\pagestyle{empty}

\begin{abstract}
  Simultaneous localization and mapping, as a fundamental task in computer vision, has gained higher demands for performance in recent years due to the rapid development of autonomous driving and unmanned aerial vehicles. Traditional SLAM algorithms highly rely on basic geometry features such as points and lines, which are susceptible to environment.  Conversely, higher-level object features offer richer information that is crucial for enhancing the overall performance of the framework. However, the effective utilization of object features necessitates careful consideration of various challenges, including complexity and process velocity. Given the advantages and disadvantages of both high-level object feature and low-level geometry features, it becomes essential to make informed choices within the SLAM framework. Taking these factors into account, this paper provides a thorough comparison between geometry features and object features, analyzes the current mainstream application methods of object features in SLAM frameworks, and presents a comprehensive overview of the main challenges involved in object-based SLAM.
\end{abstract}
\section{INTRODUCTION}
\label{INTRODUCTION}

Simultaneous Localization and Mapping is a fundamental task in computer vision, which has been highly regarded by researchers since its inception. It is defined as estimating the pose of sensors and simultaneously constructing the map of unknown environment based on observations. Many visual SLAM algorithms choose to use geometry features such as points \cite{mur2015orb, mur2017orb, campos2021orb} and lines\cite{pumarola2017pl} for pose estimation, due to its wide applicability. These approaches are commonly known as feature-based methods.

Current feature-based frameworks mostly follow a pattern of “extraction-matching-optimization”. Different types of features are extracted and associated at the front-end, while the resulting constraint relationships are optimized locally and globally at the back-end. Traditional geometry features have been an essential component in various frameworks due to their simplicity and efficiency. However, they tend to be vulnerable in scenarios with significant disturbances such as illumination and occlusion. The concept of incorporating objects as a class of features into algorithms has long been present. Unfortunately, its implementation has been hindered by limitations in accuracy and extraction speed. Nevertheless, the advancements in deep learning and computational power have considerably improved the performance of deep learning models. This has opened the possibility of real-time feature extraction using models. Consequently, in recent years, there has been a surge in research focused on object SLAM, leveraging the potential of object features.

\begin{figure}[ht]
  \centering
  \includegraphics[width=0.4\textwidth]{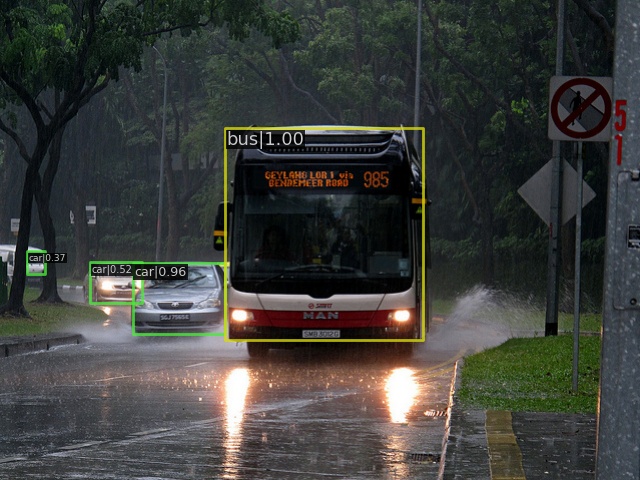}
  \caption{Object Feature in the image. Using deep learning models, information of objects can be represented by bounding box. The rebuild of objects are all depend on the result of object detection.}
  \label{object-2d}
\end{figure}
Object features is based on object detection. It can describe the distribution of objects in a scene, including their categories and spatial occupancy,as shown in Fig .\ref{object-2d}. Compared to geometry features, object features, as a higher-level category, offer the following advantages:
\begin{itemize}
  \item Richer information: Object contains far more information than geometry feature, such as semantics, shapes, and positions, which are all can be used as constraints in optimization process.
  \item Stabler observations: Objects can be observed more robustly in the scene, thus establishing more observation relationships across frames.
  \item More human-like: Perception and action planning at the object level align more with human cognitive patterns.
\end{itemize}

It is evident that object features operate in a higher dimensional space, providing richer environment information and more stable constraints, which are all crucial factors for mapping and pose estimation. However, treating objects as a distinct feature also imposes higher requirements on tasks such as parameterization and data association.

We notice that while there have been abundant summaries of geometry feature based SLAM algorithm, survey of object based SLAM is still limited. Therefore, our objective is to fill this gap by offering valuable insights to researchers interested in this field. In this paper, we will analyze the character of object features and introduce the challenges associated with their application in SLAM. By doing so, we aim to provide assistance and guidance to researchers exploring the potential of object features in SLAM.
\section{Geometry Feature and Object Feature}
\label{Geometry Feature and Object Feature}

Features can be understood as distinct elements that exhibit stability and prominence. As features, they need to satisfy two main criteria: repeatability, meaning they can appear repeatedly in different images, and distinctiveness, meaning different features should have different representations while similar features exhibit similar representations. By extracting features, we can effectively simplify the input by discarding less significant parts and focusing more on the feature elements. Furthermore, by establishing the connections between features across images, we can establish matching relationships between features and estimate the sensor's trajectory. In this section, we will primarily introduce traditional geometry features and higher-level object features.

\subsection{Geometry Feature}

Traditional geometry features in SLAM predominantly focus on extracting information from fundamental geometry elements. These features can be categorized into three main types: point features, line features, and plane features.

For these three types of geometry features, there has been extensive research. As the most fundamental geometry element, point features are the most abundant and easily extracted features in an image. This means that point features can quickly provide constraint information between frames. Typical feature points such as ORB\cite{rublee2011orb}, SIFT\cite{lowe2004distinctive}, SURF\cite{bay2008speeded} and FAST\cite{rosten2006machine} are widely used. Feature points are identified based on their distinctiveness from the surrounding points within a certain range. These points should exhibit noticeable differences in their characteristics, such as intensity, texture, or depth, compared to their neighboring points. Taking ORB feature point as an example, a point can be considered as a feature point only if the difference between the value of $k$ consecutive pixels on a circle with radius $r$ and the value of the target pixel is greater than the threshold.

Line features, building upon points, add the character of extension. The extraction of line segments is relatively more complex, evolving from early methods like Canny\cite{canny1986computational} and Hough Transformation\cite{duda1972use} to more widely used methods like LSD\cite{von2012lsd}, EDLINE\cite{akinlar2011edlines}, and ELSED\cite{suarez2022elsed}. Line extraction methods have seen significant improvements in terms of speed and accuracy. However, the underlying gradient-based extraction strategy remains unchanged. By comparing pixel gradients, it is determined whether adjacent elements belong to the same line segment. Line features, due to their elongated nature, are more susceptible to discontinuities in challenging conditions such as variations in lighting or camera shake. These factors can cause the appearance of lines to become fragmented or disrupted, making their detection and tracking more challenging in such scenarios. Additionally, 3D reconstruction of line segments is also a challenging problem. Algorithms that utilize monocular cameras, such as \cite{he2018pl} and \cite{wang2021line}, can employ methods like back-projecting the line segments onto the imaging plane for triangulation. For RGB-D cameras that directly provide depth information, methods like direct depth recovery through 3D point fitting can be used, as demonstrated in \cite{li2021rgb}. However, the depth recovery for line features is more challenging compared to the direct triangulation methods used for point features.

When the level of features rises to plane features, the methods for extraction, representation, and utilization become even more complex. \cite{shu2022structure} utilizes real-time triangulation of sparse and noisy 3D points to reconstruct 3D planes. It first predicts the plane membership of points using PlaneRecNet\cite{xie2021planerecnet} and then performs fitting using RANSAC. For algorithms using RGB-D cameras, since depth information is available, plane extraction can be achieved by leveraging the depth distribution and normal vector information of the scene, as demonstrated in \cite{holz2012real, silberman2012indoor, feng2014fast}. Unlike the methods used for point and line features that construct descriptors, plane features themselves are significantly fewer in quantity, allowing for simple identification using normal vectors and distances\cite{shu2022structure}. At the application level, unlike the establishment of observation constraints through point and line feature matching, the Manhattan World hypothesis is a popular application for plane features. It calculates the scene's Manhattan coordinate system by extracting plane normals and combining them with line segment direction vectors. Finally, constraints are established based on the parallel and perpendicular relationships between the coordinate system and line segments.

\subsection{Object Feature}

As the level of features continues to rise to the object level, it transcends the limitations imposed by geometry. Object features actually consist of two parts of information: the spatial occupancy and category of objects. Therefore, object features represent a combination of geometry and semantic features, which better align with human perception of the environment.

Semantic information of objects are generally directly extracted through well-trained deep learning models such as YOLO\cite{redmon2018yolov3} and Mask-RCNN\cite{he2017mask} during object detection. Hence, here we focus on discussing geometry representation methods for objects. When describing objects, we mainly aim for three objectives: comprehensiveness (describing objects of all categories), completeness (describing the entire content of objects), and detail (describing the fine-grained features of objects).

The earliest object descriptions can be traced back to \cite{salas2013slam++}, where pre-built object models were used for object representation. However, due to the time cost of search and the complexity of model database construction, the accuracy and richness of object representation were severely limited. \cite{joshi2018integrating} employed a categorical description approach for objects and combined it with internal line features for object matching. \cite{wang2021dsp} utilized DeepSDF\cite{park2019deepsdf} as a prior shape embedding to reconstruct objects by minimizing surface consistency and SDF Render terms. Essentially, these methods of object description are built upon rich prior knowledge. By matching the detected objects with existing knowledge, object reconstruction is aided by the prior knowledge associated with the objects. These methods achieve comprehensive and detailed object descriptions during the construction of the object database. However, the object categories are limited to the objects present in the database, which means comprehensive descriptions cannot be achieved.

\begin{figure*}[ht]
  \centering
  \includegraphics[width=\linewidth]{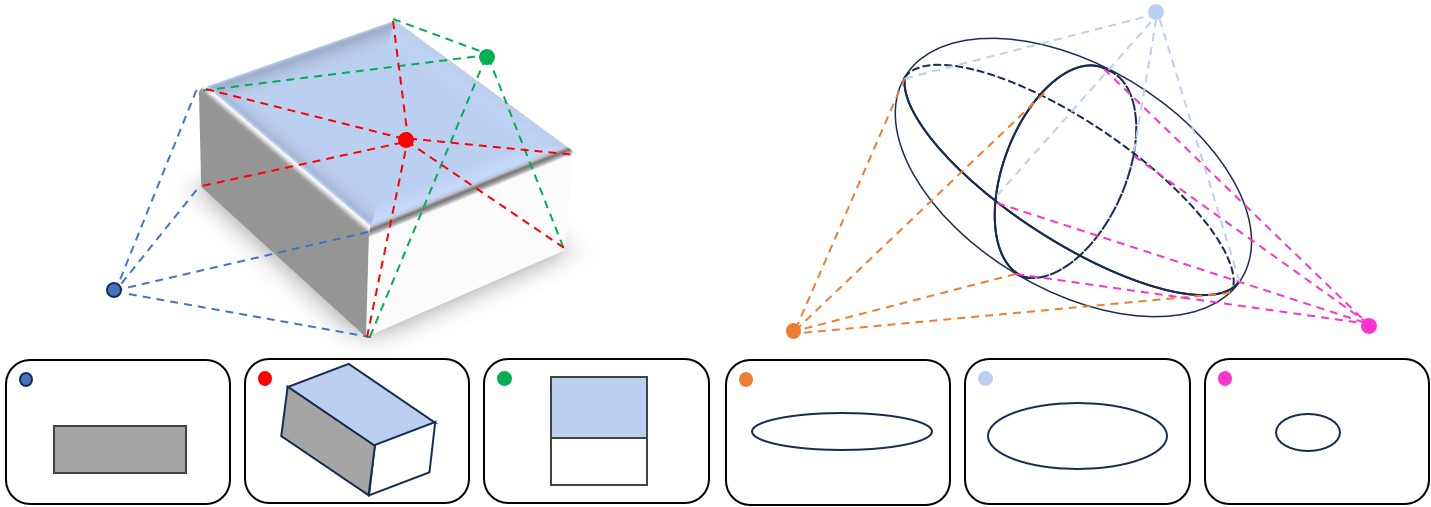}
  \caption{Difference of cube and elliposid in projecton. Under various viewing angles, the shape of the cube's projection undergoes changes, whereas the ellipsoid consistently maintains its conic nature. This characteristic leads to the requirement of categorizing cubes based on the view when calculating the reprojection error, whereas ellipsoids can be directly represented using a unified dual-conic expression.}
  \label{Difference of Projection}
\end{figure*}

To more broadly utilize objects existing in the scene, we need to employ geometry representation methods to break free from the limitations of object databases and construct objects in real-time based on observation results. One approach is to use standard geometry models for object representation. \cite{yang2019cubeslam} utilizes vanishing points(VP) from different viewing directions to recover the 3D bounding boxes of objects from 2D bounding boxes. The scheme of using bounding box is basic and intuitive for humans. However, the primary challenge of 3D bounding boxes lies in the uncertainty of projection, as different perspectives can yield varying observations. Consequently, \cite{yang2019cubeslam} introduces several assumptions to streamline real-world scenarios. Building upon this, \cite{nicholson2018quadricslam, liao2022so, ok2019robust} employ ellipsoids to describe objects. Compared to cubes, ellipsoids offer nine degrees of freedom, encompassing position (center coordinates), orientation (roll, pitch, and yaw), and shape (semi-axes in three directions). Although more constraints are necessary during initialization, as depicted in Fig .\ref{Difference of Projection}, it is evident that while the projection of a cube can exhibit significant variations based on the viewpoint, the projection of an ellipsoid consistently retains a conic shape. This characteristic proves advantageous in calculating reprojection errors and object associations.

In the process of constructing standard geometry objects, although we achieve complete construction of objects for all categories, we lose details of objects. This leads to a greater reliance on the properties of standard geometry objects in subsequent data association modules. To address this issue, some researchers have attempted more precise object description methods: directly constructing the entire irregular objects based on observation results. \cite{runz2018maskfusion} and \cite{mccormac2018fusion++}, perform scene segmentation using MaskRCNN\cite{he2017mask} and conduct online reconstruction of segmented objects. However, for segmentation tasks, only parts of  objects are visible, making it difficult to achieve a complete object description. Therefore, such object description methods are mainly used for mapping purposes.

\subsection{Problems caused by Higher Level Feature}

The process from point features to object features is actually a continuous elevation of feature hierarchy. Throughout this process, the information embedded within the features progressively amplifies, while the semantic significance of the features becomes more expressive. Moreover, as high-dimensional features encompass a larger space within the scene, they tend to maintain more stable observational relationships. These are the advantages of high-level features. However, the difficulty of feature extraction and description also increases.

\begin{figure}[ht]
  \centering
  \includegraphics[width=\linewidth]{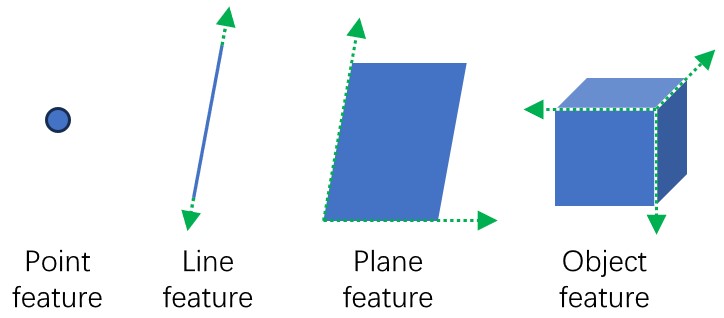}
  \caption{Illustration of feature uncertainty. As the level increases, direction of feature extension also grow, which leads to the increase of object uncertainty.}
  \label{2.jpg}
\end{figure}

We posit that the rise in difficulty primarily arises from the escalation of uncertainty induced by the expansion of features, as shown in Fig .\ref{2.jpg}. Point features, being the fundamental elements of an image, do not possess an inherent tendency for extension. However, when point features evolve into line features, the extraction process naturally introduces increased uncertainty due to the extension of these line features along their respective directions. This extension further gives rise to challenges associated with occlusion, changes in viewpoint, and resulting discontinuities. As the feature hierarchy progresses towards plane features, uncertainty arises in two directions, intensifying the unpredictability of plane features. Finally, as the highest level of features, object features exhibit three directions of uncertainty solely based on their shape, even without considering semantic information.

\section{Application of Object Feature}
\label{Application}

SLAM is evaluated based on two aspects: trajectory accuracy and constructed map. Correspondingly, features also play a dual role: connecting adjacent frames in pose estimation and serving as fundamental elements in the map. Compared to the limited applications of traditional geometry features, the richer information offered by object features has further expanded the possibilities of their application. In this section, we will primarily focus on analyzing how object features are applied in the SLAM framework.

\subsection{Object based Global Optimization}

From the perspective of trajectory accuracy, low-level geometry features such as point and line features are widely used in SLAM. After extraction and matching, geometry features are then involved in pose estimation by minimizing geometric errors \cite{mur2015orb}\cite{pumarola2017pl}. Reconstructed geometry features also play a role in global optimization and loop detection. However, when it comes to high-level object features, the extraction and representation of objects themselves incur significant computation costs. Consequently, it becomes challenging for object features to directly participate in the front-end pose estimation process as effectively as point and line features do.

Nevertheless, objects, as stable features, provide more stable observations in the scene. In the back-end optimization, where time constraints are less stringent, object features find wider application space. For example, \cite{yang2019cubeslam} optimizes the nine degrees of freedom bounding boxes of objects together with camera poses and map elements. It introduces additional measurement error terms to bundle adjustment optimization by considering the relationships between objects, poses, and map points. \cite{nicholson2018quadricslam} incorporates object observations in the form of ellipsoids as part of the factor graph and uses the deviation between the observed ellipsoids and the modeled ones as geometry error terms. Similarly, \cite{sunderhauf2017dual} performs back projection of object bounding box edges and optimizes poses and map landmarks by minimizing reprojection errors. \cite{frost2018recovering} models both points and objects as fixed-radius spheres and jointly optimizes poses.

\begin{table}[!ht]
    \centering
    \caption{AVERAGE LOCALIZATION ERROR ON TUM SEQUENCES}
    \label{accuracy}
    \begin{threeparttable} 
    \resizebox{\linewidth}{!}{
        \begin{tabular}{|c|c|c|c|c|}
        \hline
            \multirow{2}{*}{Sequence} &\multirow{2}{*}{ORB-SLAM2} &\multicolumn{2}{c|}{QuadricSLAM\tnote{*}}          &\multirow{2}{*}{Fusion++}  \\ \cline{3-4}
            \multirow{2}{*}{}         &\multirow{2}{*}{}          &Fovis\cite{huang2017visual}&ORB-VO\cite{mur2017orb}    &\multirow{2}{*}{}          \\ \hline
            \textbf{fr1\_desk} & {\color[HTML]{F8A102} \textbf{0.0159}} & 0.0632 & 0.0167 & 0.049 \\ \hline
            \textbf{fr1\_desk2} & 0.0243 & 0.0662 & 0.0245 & {\color[HTML]{F8A102} \textbf{0.0153}} \\ \hline
            \textbf{fr2\_desk} & {\color[HTML]{F8A102} \textbf{0.0087}} & 0.0568 & 0.0124 & 0.114 \\ \hline
            \textbf{fr3\_office} & {\color[HTML]{F8A102} \textbf{0.0107}} & 0.0765 & 0.0230 & 0.108 \\ \hline
        \end{tabular}
    }
    \begin{tablenotes}
    \footnotesize
    \item[*] QuadricSLAM provided results based on two sources of visual odometry.  
    \end{tablenotes}
    \end{threeparttable} 
\end{table}

Table \ref{accuracy} presents a comparison of the trajectory accuracy between object SLAM and geometry feature SLAM methods. Indeed, the introduction of object features in pose estimation has not yielded the anticipated improvements. This can be primarily attributed to the inherent complexity associated with object feature. The process of extracting meaningful object features from sensor data and accurately parameterizing them in a way that facilitates effective pose estimation is a challenging task. Additionally, the limited quantity and low reliability of object feature further hurdles in achieving accurate and robust pose estimation. Consequently, the expected benefits of object features in pose estimation have not been fully realized.

\subsection{Object based Loop Detection}
Loop detection is a crucial part in SLAM, typically employed to eliminate accumulated errors in the pose estimation process by re-observing landmarks. Traditional visual SLAM methods, such as \cite{mur2017orb}, utilize geometry features and construct bag-of-words models\cite{galvez2012bags} for loop closure detection or relocalization. However, in scenarios where there are substantial variations in lighting conditions or viewing angles, the effectiveness of traditional methods declines significantly due to the limited observability of local features. In light of this, utilizing object features, which offer more stable observations in the scene, holds great promise for loop detection.

The most commonly employed approach is to describe the scene by emphasizing the relationships between objects. This involves transforming objects in the scene into a topological graph, where points represent objects, and edges represent the relationships between objects. By prioritizing the relationships between objects, a higher-level description of the scene can be effectively achieved. For example, \cite{kong2020semantic} represents objects in the environment using a graph structure and employs graph similarity networks for recognizing repeated scenes. \cite{qian2022towards} and \cite{yu2022semanticloop} extend the concept of co-visibility to the object level and enhance the distinctiveness of nodes in the topological graph by incorporating appearance similarity and embeddings.

One limitation of these description methods is that when multiple instances of the same object category appear in the scene, relying solely on semantic features for the nodes in the topological graph can lead to a decrease in the effectiveness due to low distinctiveness. Therefore, ongoing research aims to enrich the content of nodes and edges in the topological graph. \cite{ji2023loop} describes objects using color histograms and object embeddings based on bounding boxes while constructing a topological structure containing spatial layout and semantic information of objects and their neighboring objects. \cite{wu2023object} represents object nodes using both geometric parameters and semantic information and utilizes random walk descriptors to describe the topological relationships between objects.

\begin{figure}[ht]
  \centering
  \includegraphics[width=\linewidth]{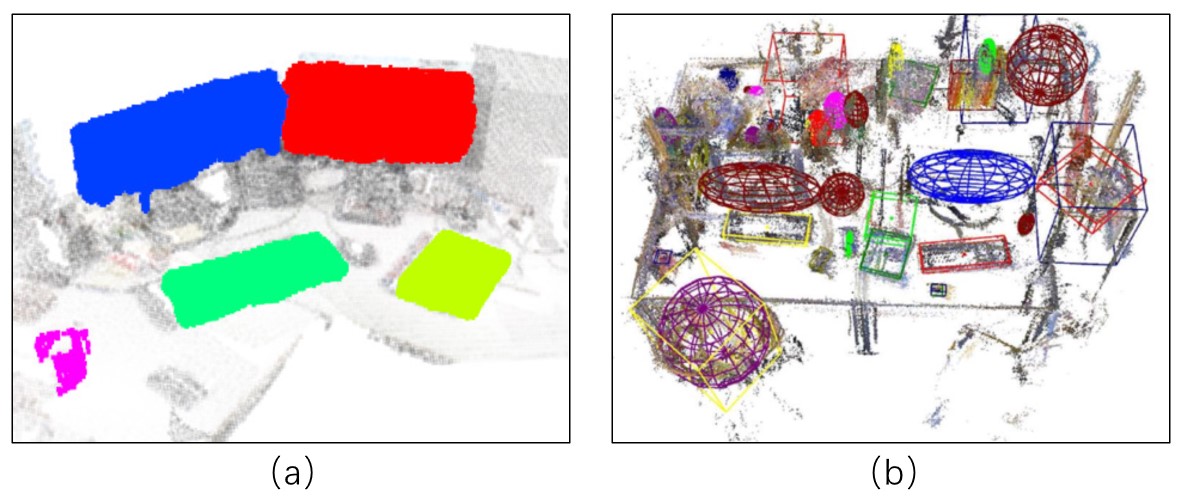}
  \caption{The result of symbolic mapping and embedded mapping. Objects are represented by different colors in \cite{sunderhauf2017meaningful}, while cubes or ellipsoid are embedded into constructed map in \cite{wu2023object}.}
  \label{map result}
\end{figure}

\subsection{Object based Mapping}
We can categorize object-level mapping based on the representation of objects into two types: symbolic and embedded. Symbolic mapping involves attaching semantic labels to elements in geometric maps, representing object information using different colors, as shown in Fig .\ref{map result}(a). Specifically, \cite{nuchter2008towards} utilizes edge of object and reflectance data to directly mark object information on the constructed map. \cite{sunderhauf2017meaningful} by incorporating 2D detection and 3D segmentation to obtain non-parametric object information, which is directly marked with different colors in the point cloud map.

On the other hand, as shown in Fig .\ref{map result}(b), embedded mapping integrates parameterized objects such as cubes or ellipsoids into the map. During the feature extraction process, it is inevitable to optimize the parameter information of objects. This parameter information includes the geometry center and shape of the objects. Therefore, we can conveniently incorporate the objects directly into the constructed map, thereby enriching the map with higher-level object information. Embedded mapping provides a higher level of map understanding compared to sparse maps and the computation cost is lower as not all points need to be processed. In the work by \cite{salas2013slam++}, objects are identified using RGB-D information, and object models from a database are inserted into the map. Similarly, \cite{nicholson2018quadricslam, liao2022so, ok2019robust, yang2019cubeslam} employ the insertion of standard geometry models like ellipsoids and cubes to represent objects within the map. This representation captures the shape and orientation of objects, enriching the map with object-level information. \cite{wu2023object} utilizes different shapes to represent maps based on object categories, creating more detailed object-level maps.

By incorporating object models into the map, these approaches enhance the understanding and representation of objects within the environment. This enables the creation of more comprehensive and detailed object-level maps, contributing to a richer understanding of the scene. From a human perspective, such maps are more aligned with human understanding habits. From a machine perspective, the inclusion of object information, particularly embedded parameterized object information, plays a facilitating role in tasks such as robotic arm grasping and 3D scene reconstruction.

\subsection{Object based High-level Application}
In the preceding sections, we summarized some application methods of object features in SLAM. It is evident that the inclusion of object features indeed leads to improvements in trajectory estimation, mapping, and other aspects. As high-level features, we believe that object features are best suited for assisting in various high-level applications.

\begin{figure}[ht]
  \centering
  \includegraphics[width=\linewidth]{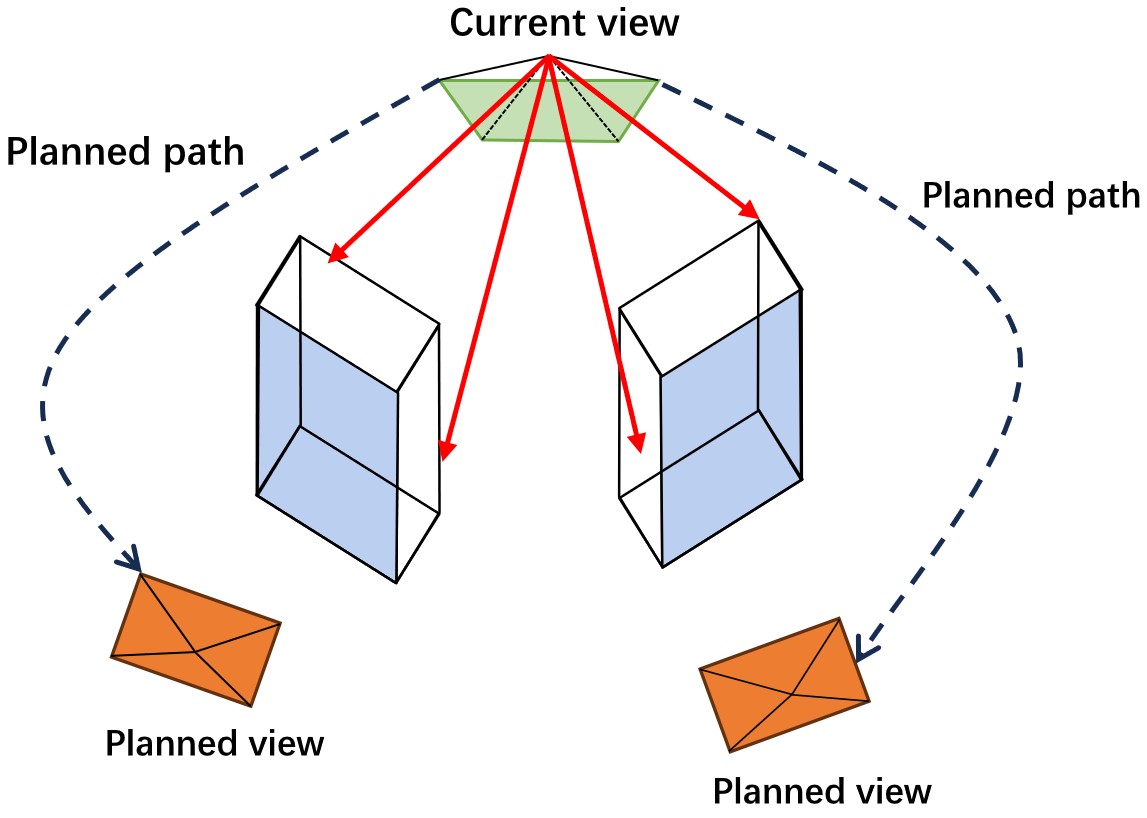}
  \caption{Illustration of object based active perception. The view and move path will be planned based on the percept status of objects in the environment.}
  \label{active perception}
\end{figure}

To achieve comprehensive environment perception, we can actively adjust the position of sensors based on the current construct state of the map to acquire information about incomplete areas. This technique is known as active perception and is commonly used in tasks such as mapping and robotic arm grasping. The key to active perception lies in enabling the robot to comprehend the incomplete parts of the map. Traditional approaches rely on global environment analysis. \cite{kahn2015active} models occluded regions using Gaussian mixture models and continuously updates them using extended Kalman filtering during the mapping process to plan sensor movements. \cite{charrow2015information} utilizes Cauchy Schwarz quadratic mutual information (CSQMI) to guide the robot's active perception of unknown areas in the map, improving the efficiency of dense 3D map construction. Although global-based approaches can plan movement routes comprehensively, they require significant computation resources to calculate the uncertainty of the entire scene. In comparison, object-based active perception is more lightweight. As shown in Fig .\ref{active perception}, \cite{wu2023object} measures the completeness of scene construction using object observation completeness. The point clouds within each object are projected onto the object's five grid surfaces, and the observation status of each surface is determined based on grid occupancy results. This information is then used to calculate the entropy of object observations, enabling the sensor to move in the direction of maximum information gain, leading to faster and more efficient environment perception. Incorporating object information can prevent mapping processes from being influenced by edge regions while reducing computational overhead and improving the overall system's efficiency.

The inclusion of object information in maps holds greater significance for practical applications as well. In the field of robotic arm grasping, visual methods efficiently construct object shapes without physical contact but may lack precision. This characteristic complements the accurate yet slower tactile-based approaches.\cite{wang20183d} utilizes visual observations to construct a 2.5D sketch, which serves as a foundation for object perception. The object models are then refined using prior knowledge of object shapes and feedback from tactile sensing, enhancing the success rate and grasp quality accuracy. \cite{sun2016object} leverages both tactile and visual sensors to identify object models, enabling efficient planning of grasp points and hand movements, resulting in faster and more accurate grasping operations. In summary, visual information provided by object perception serves as a secondary source of information alongside tactile sensing. By analyzing the geometric features of objects in advance, robots can determine preliminary grasp poses and efficiently plan suitable grasping strategies during subsequent tactile exploration processes.

\section{Challenges in Object based SLAM}
From current research, it is apparent that object features hold immense potential. However, due to their complexity, the application of object features in the SLAM framework introduces new challenges that were not present in traditional geometry SLAM approaches. To fully exploit the capabilities of object features, it is imperative for us to undertake further research dedicated to understanding and addressing these challenges.

\subsection{Object Extraction}
If we want to incorporate objects as a type of feature into the SLAM framework, accurately and stably extraction of objects from the scene becomes a crucial issue. Unlike extracting basic geometry elements, the extraction and description of objects themselves are computationally expensive tasks. The extraction process generally relies on object detection tasks, using object detection networks such as \cite{redmon2018yolov3} and \cite{he2017mask} to compute the bounding box of objects. 

Regarding the description part, different object representation methods have different compute approaches. Here, we mainly consider two types of object representation methods: ellipsoids and cuboids. In the parameterization process, these parameters follow the assumption that the object is tangent to the back projection of 2D bounding box. This assumption provides four constraints for parameter estimation, and the differences between representation methods are reflected in the remaining constraints.

For the standard cuboid object representation method, \cite{yang2019cubeslam} simplifies the three-dimensional orientation information of objects into a one-dimensional yaw angle using the ground object assumption. It calculates the position of the 3D bounding box based on the geometry relationship between vanishing points and 2D corners of the cube. Multiple bounding boxes are computed, and they are scored based on distance, angle alignment, and shape. The best bounding box is selected as the final choice. In \cite{wu2023object} an improved Isolation Forest algorithm (iForest) is used to remove outliers from the reconstructed point cloud. The retained point cloud is used to calculate scale and position information, while rotation information is estimated based on the angle between line segments and object edges. This method enables fast object initialization using a single-frame geometry element extraction approach.

On the other hand, for ellipsoids with less regular shapes, the description process is relatively more complex. \cite{nicholson2018quadricslam} utilizes the properties of dual quadrics and calculates the parameters of the ellipsoid by back projecting the 2D bounding box into 3D space. However, this method considers only this constraint, so at least three frames are required to complete the initialization of the ellipsoid. To achieve single-frame initialization, the most common approach is to introduce additional constraints. \cite{ok2019robust} introduces a texture plane measurement model and a semantic shape prior. The former states that the planes fitted to the reconstructed 3D feature points should be tangent to the object, while the latter provides prior information on scale based on the object category. \cite{liao2022so} additionally incorporates the plane supporting constraint, using the relationship between objects and the spatial structure in the environment, which is generated as objects overcome gravity, for initialization.

It can be seen that compared to basic geometry elements, object extraction and parameterization are highly complex tasks. It is necessary to use suitable methods to balance extraction speed and accurately describe objects. This greatly increases the difficulty of incorporating object features into the front-end framework.

\subsection{Data Association}
Data association, emphasizing the matching relationships between new and old features, presents another challenge in object based SLAM. While traditional geometry features can be matched using descriptor-based methods, it is difficult to directly establish matching relationships between object features due to their complexity. In order to simplify the problem, certain research approaches assume accurate data association. However, it is important to note that this assumption may potentially compromise the robustness of the framework. Hence, it becomes crucial to explore methods for establishing fast and accurate associations between object features.

\subsubsection{Strong Connection}
There are two main approaches to address the data association problem. One approach is to establish direct strong connections between objects, where the established matching relationships remain fixed throughout the subsequent processes. Since object boundaries can be representative, many studies attempt to use bounding boxes for matching. For example, \cite{li2019semantic} projects object bounding boxes and uses the Hungarian algorithm for optimal matching. Similarly, \cite{mccormac2018fusion++, wang2021dsp, sucar2020nodeslam} establish map-object matching relationships based on the intersection over union (IoU) calculation of projected object bounding boxes. This IoU calculation serves as a measure of overlap between the projected bounding boxes, enabling robust map-object matching.

In addition to bounding boxes, object features are inherently built upon three geometry features, allowing us to establish strong associations using internal geometry feature. In \cite{yang2019cubeslam}, attribution relationship are established based on the positions of feature points. A point will be divided into a object if it falls within the 2d bounding box. If detected object in different frames share a sufficient number of matching feature points, they are considered as matching objects. Similarly, \cite{sunderhauf2017meaningful} first identifies candidate near the center of a given object and then calculates the distance between objects using internal points, similar to the iterative closest point (ICP) algorithm. If the distance is below a threshold, they are considered the same object. Moreover, distribution of features within objects can also be used for matching. \cite{iqbal2018localization} and \cite{wu2023object} consider the probability distribution of internal feature points and use nonparametric statistics to establish stable connections between objects. 

\begin{figure}[ht]
  \centering
  \includegraphics[width=\linewidth]{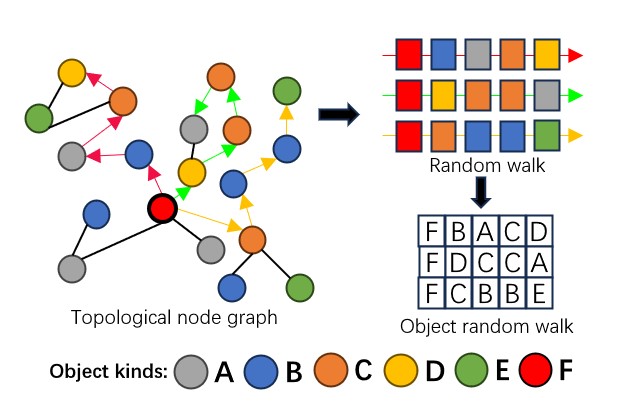}
  \caption{Illustration of random walk descriptor. After objects in scene convert to topological diagram, the interrelationship among nodes can be used to describe objects. The richer the information encapsulated by a node, the greater the uniqueness of its descriptors.}
  \label{object random walk}
\end{figure}

Except objects themselves, the relationships between objects can also be used to build association. \cite{liu2019global} treat detected objects as nodes and introduces random walk descriptors that leverage the topological relationships between objects for data association. This methodology, illustrated in Fig. \ref{object random walk}, allows for utilizing the object relationships to improve the accuracy of data association. Based on this idea, \cite{wu2023object} supplements object parameterization information to the node's descriptor, enhancing the accuracy of data association.

\subsubsection{Weak Connection}
Algorithm mention above explore various strategies to establish strong associations between object features. But mismatch is inevitable, compare with spending a lot of effort to establish accurate matching relationships, why can't we think of matching as dynamic problem. The relationships established at a specific moment may be the most likely judgments based on the observation results at that moment. There may exist mismatches but it can be corrected with the execution of algorithm. The matching relationships established based on this idea are referred to weak connections.

\cite{bowman2017probabilistic} and \cite{doherty2020probabilistic} construct probability models for the matching relationships and optimize them using the Gauss-Newton method. \cite{zhang2019hierarchical} models the association relationships between objects using a hierarchical Dirichlet model. This weak connection approach transforms the matching relationships into a probability problem, known as probabilistic data association, gradually maximizing the overall probability during the map construction process.

Compared to strong connections, this method allows for dynamic adjustment of the matching relationships between objects during the optimization process. It exhibits better tolerance towards mismatches. However, a major challenge is the significant time overhead in the optimization process, which can be particularly problematic for object SLAM frameworks that already have high time costs.

\subsection{Dataset and Evaluation}
To accurately assess the performance of a SLAM algorithm, it is essential to establish appropriate evaluation criteria. These criteria serve as benchmarks for measuring the algorithm's effectiveness and provide insights into its capabilities. Table \ref{Evaluation Criteria} presents the current mainstream evaluation criteria for object SLAM. Almost every algorithm uses different evaluation criteria. To better measure performance of algorithm, there is a need for universally standardized and objective evaluation criteria that align more closely with the characteristics of objects.

\begin{table}[ht]
  \centering
  \caption{EVALUATION CRITERIA}
  \label{Evaluation Criteria} 
  \begin{threeparttable} 
  {
    \begin{tabularx}{\linewidth} {
        | >{\hsize=.5\hsize\linewidth=\hsize}X
        | >{\hsize=1.5\hsize\linewidth=\hsize}X
        |}
      \hline
            Algorithm\tnote{*} & Evaluation Criteria \\ \hline
            Wu\cite{wu2023object} & Accuracy of data association and object extraction, amount of object landmark, success rate and time cost of scene matching. \\ \hline
            Niko\cite{sunderhauf2017meaningful} & Result of semantic mapping. \\ \hline
            Fusion++\cite{mccormac2018fusion++} & Accuracy of loop detection, RMSE of absolute trajectory error, result of rebuilt object. \\ \hline
            CubeSLAM\cite{yang2019cubeslam} & RMSE of absolute trajectory error, accuracy of objection extraction, result of dynamic object detection.  \\ \hline
            QuadricSLAM\cite{nicholson2018quadricslam} & RMSE of absolute trajectory error, result of semantic mapping, amount of extracted object. \\ \hline
            ROSHAN\cite{ok2019robust} & Result of semantic mapping, amount of extracted object. \\ \hline
            SO-SLAM\cite{liao2022so} & Accuracy of objection extraction, time cost. \\ \hline
    \end{tabularx}
  }
  \begin{tablenotes}
    \footnotesize
    \item[*] For algorithm without abbreviation, we use author's last name as alternative.
   \end{tablenotes}
   \end{threeparttable} 
\end{table}

Furthermore, we have observed that there is a lack of object-level annotations in current datasets when evaluating object extraction using object IoU. Although datasets like TUM\cite{sturm2012benchmark} and KITTI\cite{geiger2012we} have provided bounding box annotations for objects, these annotations are limited to representing objects as standard cuboids. It becomes challenging to accurately measure the accuracy of object-level mapping for more diverse object representations, such as ellipsoids. This issue highlights the need for addressing dataset limitations in object based SLAM.

\section{Conclusions}
In this paper, we analyze the main character of object feature and compare it with traditional geometry feature. Based on the property of object feature, we introduce application methods of object feature and summarize challenges of utilizing object feature in SLAM framework. Although the existing methods have limited effectiveness in improving trajectory accuracy, we found that object features have broad application space in mapping and high-level applications. In our future work, we will continue to explore the application  of object feature, and strive to develop a more comprehensive object SLAM framework.

\section*{ACKNOWLEDGMENT}
This work was partially supported by the Gansu Provincial Sci ence and Technology Major Special Innovation Consortium Project (No. 21ZD3GA002), the Fundamental Research Funds for the Central Universities under Grant No. lzujbky\-2022\-kb12, lzujbky\-2021\-sp43, Science and Technology Plan of Qinghai Province under Grant No.2020-GX-164, Google Research Awards and Google Faculty Award. We also gratefully acknowledge the support of NVIDIA Corporation with the donation of the Jetson TX1 used for this research.

\addtolength{\textheight}{-6cm}

\bibliographystyle{IEEEtran}
\bibliography{IEEEabrv,IEEEexample}

\begin{thebibliography}{10}
\providecommand{\url}[1]{#1}
\csname url@rmstyle\endcsname
\providecommand{\newblock}{\relax}
\providecommand{\bibinfo}[2]{#2}
\providecommand\BIBentrySTDinterwordspacing{\spaceskip=0pt\relax}
\providecommand\BIBentryALTinterwordstretchfactor{4}
\providecommand\BIBentryALTinterwordspacing{\spaceskip=\fontdimen2\font plus
\BIBentryALTinterwordstretchfactor\fontdimen3\font minus \fontdimen4\font\relax}
\providecommand\BIBforeignlanguage[2]{{%
\expandafter\ifx\csname l@#1\endcsname\relax
\typeout{** WARNING: IEEEtran.bst: No hyphenation pattern has been}%
\typeout{** loaded for the language `#1'. Using the pattern for}%
\typeout{** the default language instead.}%
\else
\language=\csname l@#1\endcsname
\fi
#2}}

\bibitem{mur2015orb}
R.~Mur-Artal, J.~M.~M. Montiel, and J.~D. Tardos, ``Orb-slam: a versatile and accurate monocular slam system,'' \emph{IEEE transactions on robotics}, vol.~31, no.~5, pp. 1147--1163, 2015.

\bibitem{mur2017orb}
R.~Mur-Artal and J.~D. Tard{\'o}s, ``Orb-slam2: An open-source slam system for monocular, stereo, and rgb-d cameras,'' \emph{IEEE transactions on robotics}, vol.~33, no.~5, pp. 1255--1262, 2017.

\bibitem{campos2021orb}
C.~Campos, R.~Elvira, J.~J.~G. Rodr{\'\i}guez, J.~M. Montiel, and J.~D. Tard{\'o}s, ``Orb-slam3: An accurate open-source library for visual, visual--inertial, and multimap slam,'' \emph{IEEE Transactions on Robotics}, vol.~37, no.~6, pp. 1874--1890, 2021.

\bibitem{pumarola2017pl}
A.~Pumarola, A.~Vakhitov, A.~Agudo, A.~Sanfeliu, and F.~Moreno-Noguer, ``Pl-slam: Real-time monocular visual slam with points and lines,'' in \emph{2017 IEEE international conference on robotics and automation (ICRA)}.\hskip 1em plus 0.5em minus 0.4em\relax IEEE, 2017, pp. 4503--4508.

\bibitem{rublee2011orb}
E.~Rublee, V.~Rabaud, K.~Konolige, and G.~Bradski, ``Orb: An efficient alternative to sift or surf,'' in \emph{2011 International conference on computer vision}.\hskip 1em plus 0.5em minus 0.4em\relax Ieee, 2011, pp. 2564--2571.

\bibitem{lowe2004distinctive}
D.~G. Lowe, ``Distinctive image features from scale-invariant keypoints,'' \emph{International journal of computer vision}, vol.~60, pp. 91--110, 2004.

\bibitem{bay2008speeded}
H.~Bay, A.~Ess, T.~Tuytelaars, and L.~Van~Gool, ``Speeded-up robust features (surf),'' \emph{Computer vision and image understanding}, vol. 110, no.~3, pp. 346--359, 2008.

\bibitem{rosten2006machine}
E.~Rosten and T.~Drummond, ``Machine learning for high-speed corner detection,'' in \emph{Computer Vision--ECCV 2006: 9th European Conference on Computer Vision, Graz, Austria, May 7-13, 2006. Proceedings, Part I 9}.\hskip 1em plus 0.5em minus 0.4em\relax Springer, 2006, pp. 430--443.

\bibitem{canny1986computational}
J.~Canny, ``A computational approach to edge detection,'' \emph{IEEE Transactions on pattern analysis and machine intelligence}, no.~6, pp. 679--698, 1986.

\bibitem{duda1972use}
R.~O. Duda and P.~E. Hart, ``Use of the hough transformation to detect lines and curves in pictures,'' \emph{Communications of the ACM}, vol.~15, no.~1, pp. 11--15, 1972.

\bibitem{von2012lsd}
R.~G. Von~Gioi, J.~Jakubowicz, J.-M. Morel, and G.~Randall, ``Lsd: A line segment detector,'' \emph{Image Processing On Line}, vol.~2, pp. 35--55, 2012.

\bibitem{akinlar2011edlines}
C.~Akinlar and C.~Topal, ``Edlines: A real-time line segment detector with a false detection control,'' \emph{Pattern Recognition Letters}, vol.~32, no.~13, pp. 1633--1642, 2011.

\bibitem{suarez2022elsed}
I.~Su{\'a}rez, J.~M. Buenaposada, and L.~Baumela, ``Elsed: Enhanced line segment drawing,'' \emph{Pattern Recognition}, vol. 127, p. 108619, 2022.

\bibitem{he2018pl}
Y.~He, J.~Zhao, Y.~Guo, W.~He, and K.~Yuan, ``Pl-vio: Tightly-coupled monocular visual--inertial odometry using point and line features,'' \emph{Sensors}, vol.~18, no.~4, p. 1159, 2018.

\bibitem{wang2021line}
Q.~Wang, Z.~Yan, J.~Wang, F.~Xue, W.~Ma, and H.~Zha, ``Line flow based simultaneous localization and mapping,'' \emph{IEEE Transactions on Robotics}, vol.~37, no.~5, pp. 1416--1432, 2021.

\bibitem{li2021rgb}
Y.~Li, R.~Yunus, N.~Brasch, N.~Navab, and F.~Tombari, ``Rgb-d slam with structural regularities,'' in \emph{2021 IEEE international conference on Robotics and automation (ICRA)}.\hskip 1em plus 0.5em minus 0.4em\relax IEEE, 2021, pp. 11\,581--11\,587.

\bibitem{shu2022structure}
F.~Shu, J.~Wang, A.~Pagani, and D.~Stricker, ``Structure plp-slam: Efficient sparse mapping and localization using point, line and plane for monocular, rgb-d and stereo cameras,'' \emph{arXiv preprint arXiv:2207.06058}, 2022.

\bibitem{xie2021planerecnet}
Y.~Xie, F.~Shu, J.~Rambach, A.~Pagani, and D.~Stricker, ``Planerecnet: multi-task learning with cross-task consistency for piece-wise plane detection and reconstruction from a single rgb image,'' \emph{arXiv preprint arXiv:2110.11219}, 2021.

\bibitem{holz2012real}
D.~Holz, S.~Holzer, R.~B. Rusu, and S.~Behnke, ``Real-time plane segmentation using rgb-d cameras,'' in \emph{RoboCup 2011: Robot Soccer World Cup XV 15}.\hskip 1em plus 0.5em minus 0.4em\relax Springer, 2012, pp. 306--317.

\bibitem{silberman2012indoor}
N.~Silberman, D.~Hoiem, P.~Kohli, and R.~Fergus, ``Indoor segmentation and support inference from rgbd images,'' in \emph{Computer Vision--ECCV 2012: 12th European Conference on Computer Vision, Florence, Italy, October 7-13, 2012, Proceedings, Part V 12}.\hskip 1em plus 0.5em minus 0.4em\relax Springer, 2012, pp. 746--760.

\bibitem{feng2014fast}
C.~Feng, Y.~Taguchi, and V.~R. Kamat, ``Fast plane extraction in organized point clouds using agglomerative hierarchical clustering,'' in \emph{2014 IEEE International Conference on Robotics and Automation (ICRA)}.\hskip 1em plus 0.5em minus 0.4em\relax IEEE, 2014, pp. 6218--6225.

\bibitem{redmon2018yolov3}
J.~Redmon and A.~Farhadi, ``Yolov3: An incremental improvement,'' \emph{arXiv preprint arXiv:1804.02767}, 2018.

\bibitem{he2017mask}
K.~He, G.~Gkioxari, P.~Doll{\'a}r, and R.~Girshick, ``Mask r-cnn,'' in \emph{Proceedings of the IEEE international conference on computer vision}, 2017, pp. 2961--2969.

\bibitem{salas2013slam++}
R.~F. Salas-Moreno, R.~A. Newcombe, H.~Strasdat, P.~H. Kelly, and A.~J. Davison, ``Slam++: Simultaneous localisation and mapping at the level of objects,'' in \emph{Proceedings of the IEEE conference on computer vision and pattern recognition}, 2013, pp. 1352--1359.

\bibitem{joshi2018integrating}
N.~Joshi, Y.~Sharma, P.~Parkhiya, R.~Khawad, K.~M. Krishna, and B.~Bhowmick, ``Integrating objects into monocular slam: Line based category specific models,'' in \emph{Proceedings of the 11th Indian Conference on Computer Vision, Graphics and Image Processing}, 2018, pp. 1--9.

\bibitem{wang2021dsp}
J.~Wang, M.~R{\"u}nz, and L.~Agapito, ``Dsp-slam: Object oriented slam with deep shape priors,'' in \emph{2021 International Conference on 3D Vision (3DV)}.\hskip 1em plus 0.5em minus 0.4em\relax IEEE, 2021, pp. 1362--1371.

\bibitem{park2019deepsdf}
J.~J. Park, P.~Florence, J.~Straub, R.~Newcombe, and S.~Lovegrove, ``Deepsdf: Learning continuous signed distance functions for shape representation,'' in \emph{Proceedings of the IEEE/CVF conference on computer vision and pattern recognition}, 2019, pp. 165--174.

\bibitem{yang2019cubeslam}
S.~Yang and S.~Scherer, ``Cubeslam: Monocular 3-d object slam,'' \emph{IEEE Transactions on Robotics}, vol.~35, no.~4, pp. 925--938, 2019.

\bibitem{nicholson2018quadricslam}
L.~Nicholson, M.~Milford, and N.~S{\"u}nderhauf, ``Quadricslam: Dual quadrics from object detections as landmarks in object-oriented slam,'' \emph{IEEE Robotics and Automation Letters}, vol.~4, no.~1, pp. 1--8, 2018.

\bibitem{liao2022so}
Z.~Liao, Y.~Hu, J.~Zhang, X.~Qi, X.~Zhang, and W.~Wang, ``So-slam: Semantic object slam with scale proportional and symmetrical texture constraints,'' \emph{IEEE Robotics and Automation Letters}, vol.~7, no.~2, pp. 4008--4015, 2022.

\bibitem{ok2019robust}
K.~Ok, K.~Liu, K.~Frey, J.~P. How, and N.~Roy, ``Robust object-based slam for high-speed autonomous navigation,'' in \emph{2019 International Conference on Robotics and Automation (ICRA)}.\hskip 1em plus 0.5em minus 0.4em\relax IEEE, 2019, pp. 669--675.

\bibitem{runz2018maskfusion}
M.~Runz, M.~Buffier, and L.~Agapito, ``Maskfusion: Real-time recognition, tracking and reconstruction of multiple moving objects,'' in \emph{2018 IEEE International Symposium on Mixed and Augmented Reality (ISMAR)}.\hskip 1em plus 0.5em minus 0.4em\relax IEEE, 2018, pp. 10--20.

\bibitem{mccormac2018fusion++}
J.~McCormac, R.~Clark, M.~Bloesch, A.~Davison, and S.~Leutenegger, ``Fusion++: Volumetric object-level slam,'' in \emph{2018 international conference on 3D vision (3DV)}.\hskip 1em plus 0.5em minus 0.4em\relax IEEE, 2018, pp. 32--41.

\bibitem{sunderhauf2017dual}
N.~S{\"u}nderhauf and M.~Milford, ``Dual quadrics from object detection boundingboxes as landmark representations in slam,'' \emph{arXiv preprint arXiv:1708.00965}, 2017.

\bibitem{frost2018recovering}
D.~Frost, V.~Prisacariu, and D.~Murray, ``Recovering stable scale in monocular slam using object-supplemented bundle adjustment,'' \emph{IEEE Transactions on Robotics}, vol.~34, no.~3, pp. 736--747, 2018.

\bibitem{huang2017visual}
A.~S. Huang, A.~Bachrach, P.~Henry, M.~Krainin, D.~Maturana, D.~Fox, and N.~Roy, ``Visual odometry and mapping for autonomous flight using an rgb-d camera,'' in \emph{Robotics Research: The 15th International Symposium ISRR}.\hskip 1em plus 0.5em minus 0.4em\relax Springer, 2017, pp. 235--252.

\bibitem{galvez2012bags}
D.~G{\'a}lvez-L{\'o}pez and J.~D. Tardos, ``Bags of binary words for fast place recognition in image sequences,'' \emph{IEEE Transactions on Robotics}, vol.~28, no.~5, pp. 1188--1197, 2012.

\bibitem{kong2020semantic}
X.~Kong, X.~Yang, G.~Zhai, X.~Zhao, X.~Zeng, M.~Wang, Y.~Liu, W.~Li, and F.~Wen, ``Semantic graph based place recognition for 3d point clouds,'' in \emph{2020 IEEE/RSJ International Conference on Intelligent Robots and Systems (IROS)}.\hskip 1em plus 0.5em minus 0.4em\relax IEEE, 2020, pp. 8216--8223.

\bibitem{qian2022towards}
Z.~Qian, J.~Fu, and J.~Xiao, ``Towards accurate loop closure detection in semantic slam with 3d semantic covisibility graphs,'' \emph{IEEE Robotics and Automation Letters}, vol.~7, no.~2, pp. 2455--2462, 2022.

\bibitem{yu2022semanticloop}
J.~Yu and S.~Shen, ``Semanticloop: loop closure with 3d semantic graph matching,'' \emph{IEEE Robotics and Automation Letters}, vol.~8, no.~2, pp. 568--575, 2022.

\bibitem{ji2023loop}
X.~Ji, P.~Liu, H.~Niu, X.~Chen, R.~Ying, and F.~Wen, ``Loop closure detection based on object-level spatial layout and semantic consistency,'' \emph{arXiv preprint arXiv:2304.05146}, 2023.

\bibitem{wu2023object}
Y.~Wu, Y.~Zhang, D.~Zhu, Z.~Deng, W.~Sun, X.~Chen, and J.~Zhang, ``An object slam framework for association, mapping, and high-level tasks,'' \emph{IEEE Transactions on Robotics}, 2023.

\bibitem{sunderhauf2017meaningful}
N.~S{\"u}nderhauf, T.~T. Pham, Y.~Latif, M.~Milford, and I.~Reid, ``Meaningful maps with object-oriented semantic mapping,'' in \emph{2017 IEEE/RSJ International Conference on Intelligent Robots and Systems (IROS)}.\hskip 1em plus 0.5em minus 0.4em\relax IEEE, 2017, pp. 5079--5085.

\bibitem{nuchter2008towards}
A.~N{\"u}chter and J.~Hertzberg, ``Towards semantic maps for mobile robots,'' \emph{Robotics and Autonomous Systems}, vol.~56, no.~11, pp. 915--926, 2008.

\bibitem{kahn2015active}
G.~Kahn, P.~Sujan, S.~Patil, S.~Bopardikar, J.~Ryde, K.~Goldberg, and P.~Abbeel, ``Active exploration using trajectory optimization for robotic grasping in the presence of occlusions,'' in \emph{2015 IEEE International Conference on Robotics and Automation (ICRA)}.\hskip 1em plus 0.5em minus 0.4em\relax IEEE, 2015, pp. 4783--4790.

\bibitem{charrow2015information}
B.~Charrow, G.~Kahn, S.~Patil, S.~Liu, K.~Goldberg, P.~Abbeel, N.~Michael, and V.~Kumar, ``Information-theoretic planning with trajectory optimization for dense 3d mapping.'' in \emph{Robotics: Science and Systems}, vol.~11.\hskip 1em plus 0.5em minus 0.4em\relax Rome, 2015, pp. 3--12.

\bibitem{wang20183d}
S.~Wang, J.~Wu, X.~Sun, W.~Yuan, W.~T. Freeman, J.~B. Tenenbaum, and E.~H. Adelson, ``3d shape perception from monocular vision, touch, and shape priors,'' in \emph{2018 IEEE/RSJ International Conference on Intelligent Robots and Systems (IROS)}.\hskip 1em plus 0.5em minus 0.4em\relax IEEE, 2018, pp. 1606--1613.

\bibitem{sun2016object}
F.~Sun, C.~Liu, W.~Huang, and J.~Zhang, ``Object classification and grasp planning using visual and tactile sensing,'' \emph{IEEE Transactions on Systems, Man, and Cybernetics: Systems}, vol.~46, no.~7, pp. 969--979, 2016.

\bibitem{li2019semantic}
J.~Li, D.~Meger, and G.~Dudek, ``Semantic mapping for view-invariant relocalization,'' in \emph{2019 International Conference on Robotics and Automation (ICRA)}.\hskip 1em plus 0.5em minus 0.4em\relax IEEE, 2019, pp. 7108--7115.

\bibitem{sucar2020nodeslam}
E.~Sucar, K.~Wada, and A.~Davison, ``Nodeslam: Neural object descriptors for multi-view shape reconstruction,'' in \emph{2020 International Conference on 3D Vision (3DV)}.\hskip 1em plus 0.5em minus 0.4em\relax IEEE, 2020, pp. 949--958.

\bibitem{iqbal2018localization}
A.~Iqbal and N.~R. Gans, ``Localization of classified objects in slam using nonparametric statistics and clustering,'' in \emph{2018 IEEE/RSJ International Conference on Intelligent Robots and Systems (IROS)}.\hskip 1em plus 0.5em minus 0.4em\relax IEEE, 2018, pp. 161--168.

\bibitem{liu2019global}
Y.~Liu, Y.~Petillot, D.~Lane, and S.~Wang, ``Global localization with object-level semantics and topology,'' in \emph{2019 International Conference on Robotics and Automation (ICRA)}.\hskip 1em plus 0.5em minus 0.4em\relax IEEE, 2019, pp. 4909--4915.

\bibitem{bowman2017probabilistic}
S.~L. Bowman, N.~Atanasov, K.~Daniilidis, and G.~J. Pappas, ``Probabilistic data association for semantic slam,'' in \emph{2017 IEEE international conference on robotics and automation (ICRA)}.\hskip 1em plus 0.5em minus 0.4em\relax IEEE, 2017, pp. 1722--1729.

\bibitem{doherty2020probabilistic}
K.~J. Doherty, D.~P. Baxter, E.~Schneeweiss, and J.~J. Leonard, ``Probabilistic data association via mixture models for robust semantic slam,'' in \emph{2020 IEEE International Conference on Robotics and Automation (ICRA)}.\hskip 1em plus 0.5em minus 0.4em\relax IEEE, 2020, pp. 1098--1104.

\bibitem{zhang2019hierarchical}
J.~Zhang, M.~Gui, Q.~Wang, R.~Liu, J.~Xu, and S.~Chen, ``Hierarchical topic model based object association for semantic slam,'' \emph{IEEE transactions on visualization and computer graphics}, vol.~25, no.~11, pp. 3052--3062, 2019.

\bibitem{sturm2012benchmark}
J.~Sturm, N.~Engelhard, F.~Endres, W.~Burgard, and D.~Cremers, ``A benchmark for the evaluation of rgb-d slam systems,'' in \emph{2012 IEEE/RSJ international conference on intelligent robots and systems}.\hskip 1em plus 0.5em minus 0.4em\relax IEEE, 2012, pp. 573--580.

\bibitem{geiger2012we}
A.~Geiger, P.~Lenz, and R.~Urtasun, ``Are we ready for autonomous driving? the kitti vision benchmark suite,'' in \emph{2012 IEEE conference on computer vision and pattern recognition}.\hskip 1em plus 0.5em minus 0.4em\relax IEEE, 2012, pp. 3354--3361.

\end{thebibliography}

\end{document}